\documentclass{article}

\usepackage{microtype}
\usepackage{graphicx}
\usepackage{subfigure}
\usepackage{booktabs} %

\usepackage{hyperref}

\usepackage[accepted]{icml2023}

\usepackage{amsmath}
\usepackage{amssymb}
\usepackage{mathtools}
\usepackage{amsthm}

\usepackage[capitalize,noabbrev]{cleveref}

\theoremstyle{plain}

\theoremstyle{definition}

\theoremstyle{remark}

\usepackage[textsize=tiny]{todonotes}

\graphicspath{{figures/}}

\icmltitlerunning{BatchGFN: Generative Flow Networks for Batch Active Learning}

\begin{document}

\twocolumn[
\icmltitle{BatchGFN: Generative Flow Networks for Batch Active Learning}

\icmlsetsymbol{equal}{*}

\begin{icmlauthorlist}
\icmlauthor{Shreshth A. Malik}{oatml}
\icmlauthor{Salem Lahlou}{mila,mont}
\icmlauthor{Andrew Jesson}{oatml}
\icmlauthor{Moksh Jain}{mila,mont}
\icmlauthor{Nikolay Malkin}{mila,mont}
\icmlauthor{Tristan Deleu}{mila,mont}
\icmlauthor{Yoshua Bengio}{mila,mont,cifar}
\icmlauthor{Yarin Gal}{oatml}

\end{icmlauthorlist}

\icmlaffiliation{mont}{Universit\'{e} de Montr\'{e}al}
\icmlaffiliation{oatml}{OATML, University of Oxford}
\icmlaffiliation{mila}{Mila – Qu\'{e}bec AI Institute}
\icmlaffiliation{cifar}{CIFAR Fellow}

\icmlcorrespondingauthor{Shreshth A. Malik}{shreshth@robots.ox.ac.uk}

\icmlkeywords{Machine Learning, ICML}

\vskip 0.3in
]

\makeatletter

\def\paragraph{\@startsection{paragraph}{4}{\z@}{0.3ex plus
   0.2ex minus .2ex}{-0.5em}{\normalsize\bf}}
  
\makeatother

\printAffiliationsAndNotice{}  %

\begin{abstract}

We introduce BatchGFN---a novel approach for pool-based active learning that uses generative flow networks to sample sets of data points proportional to a batch reward. 
With an appropriate reward function to quantify the utility of acquiring a batch, such as
the joint mutual information between the batch and the model parameters, BatchGFN is able to construct highly informative batches for active learning in a principled way. 
We show our approach enables sampling near-optimal utility batches at inference time with a single forward pass per point in the batch in toy regression problems. This alleviates the computational complexity of batch-aware algorithms and removes the need for greedy approximations to find maximizers for the batch reward.
We also present early results for amortizing training across acquisition steps, which will enable scaling to real-world tasks.

\end{abstract}

\begin{figure}
    \centering
    \includegraphics[width=\linewidth,trim=0 20 0 0]{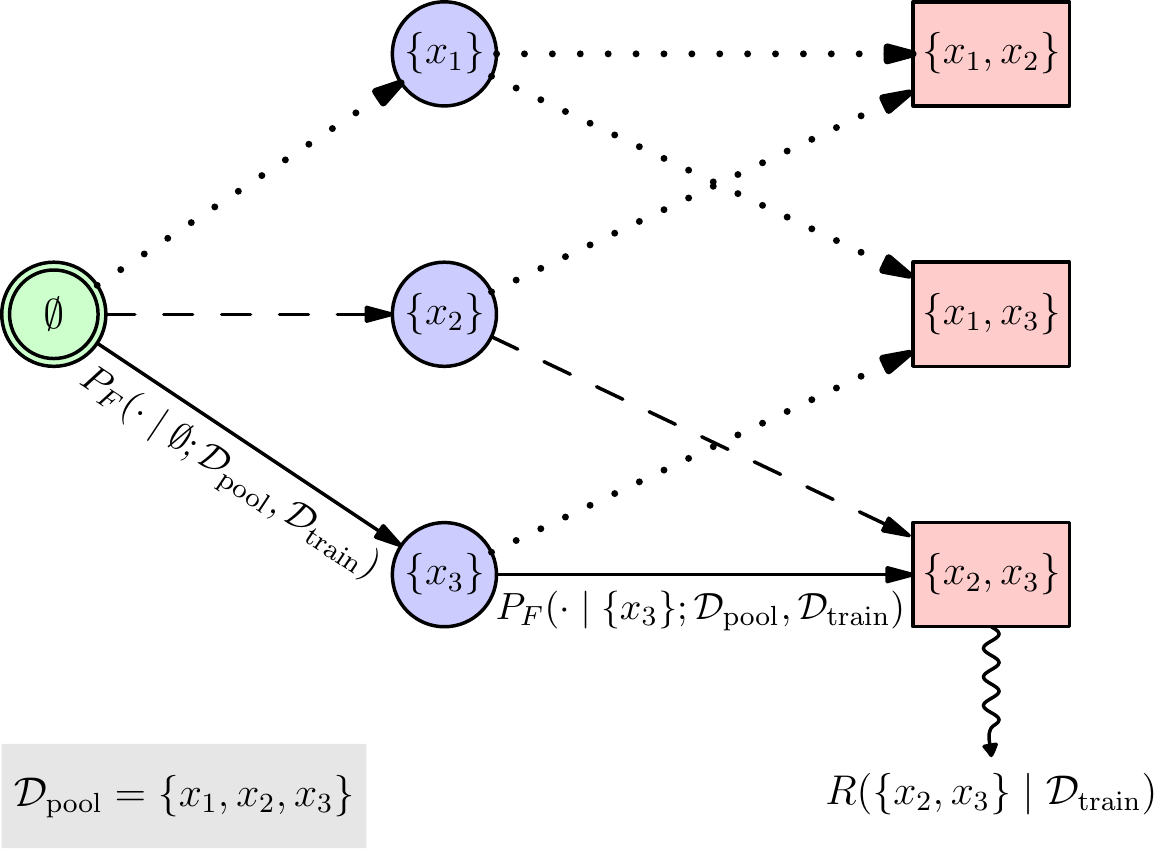}
     \caption{BatchGFN state space for constructing a batch of size 2 from a pool set of 3 points. Starting with an empty set, the policy sequentially chooses points to add to batch until a \fbox{terminal state}, representing a complete batch, is reached. Notice that a state may be reachable by more than one trajectory (e.g., the dashed path). 
     }
    \label{fig:BatchGFN}
\end{figure}

\section{Introduction}
\label{sec:intro}

Annotating large quantities of data can be prohibitively expensive, particularly in domains which require expertise. 
Active learning (AL) seeks to overcome this labelling bottleneck by iteratively selecting the most useful points to label to improve model performance \cite{houlsby2011bayesian, settles2009active}. 
In the batch setting, we seek to choose a \textit{set} of points (the query batch) from an unlabelled pool at each acquisition step. Approaches based on single-point scores tend to have redundancy in information in the batch \cite{gal2017deep}. Stochastic schemes \cite{kirsch2021stochastic} can alleviate some of this redundancy but do not have guarantees for optimal batch formation. However, more principled batch-aware methods are in general computationally expensive and often resort to greedy approximations to their true objective to construct the batch \cite{ash2021gone, ash2019deep, holzmuller2022framework, kirsch2019batchbald, seneractive}. 

In this work, we propose to select batches using a parameterized
sampler trained to sample batches proportional to a batch reward function. We use a generative flow network \cite{bengio2021flow, bengio2021gflownet} where the state of the network is the query batch under construction, and actions add pool data points to the batch. Once trained, we show that our approach enables sampling highly informative query batches more efficiently than other batch-aware algorithms, without using greedy approximations.

\section{Background}
\label{sec:bg}

\subsection{Batch Active Learning}
\label{sec:batch_al}

The goal of AL is to train models with as little data as possible. 
In the pool-based batch AL framework \cite{lewis1995sequential, settles2009active}, at each data acquisition step, we seek to choose $B$ points from a pool set $\mathcal{D}_\text{pool}=\{x_i\}_{i=1}^{N}$ to be labelled by an oracle and added to the training set $\mathcal{D}_{\text{train}}=\{x_i, y_i\}_{i=1}^{M}$, where $x \in \mathcal{X}$ are the data features and $y \in \mathcal{Y}$ are the corresponding labels. We can formalize batch AL strategies by defining an \textit{acquisition} or \textit{reward} function $R: \mathcal{X}^B \rightarrow \mathbb{R}$ which scores each potential batch. As the number of batches to score grows exponentially with $|\mathcal{D}_{\text{pool}}|$, the maximization problem is intractable, so greedy approximations are often used.

For example, the BALD \cite{gal2017deep,houlsby2011bayesian} algorithm takes a Bayesian perspective for a model with parameters $\theta$ and selects points that maximize the mutual information (MI) between the model predictions and its parameters, $\mathbb{I}\left[y, \theta \mid x, \mathcal{D}_{\text {train }}\right] = \mathbb{H}[y \mid x, \mathcal{D}_{\text {train }}] - \mathbb{E}_{\theta \sim p\left(\theta \mid \mathcal{D}_{\text {train }}\right)} [\mathbb{H}[y \mid x, \theta]]$, where $\mathbb{H}$ denotes the entropy. Intuitively, labelling points with high mutual information will decrease the uncertainty in the model parameters. The batch reward for BALD is simply the sum of the individual scores for each data point $R_{\text{BALD}} = \sum_{i=1}^B \mathbb{I}\left[y_i, \theta \mid x_i, \mathcal{D}\right]$, so the top $B$ scoring points are greedily selected. 
BALD has been shown to be ineffective to acquire batches, given that the maximizers of MI are usually similar. Injecting noise into top-$B$ acquisition scores such as BALD can be used to induce diversity in the batch, and can be viewed as approximating future acquisition scores \cite{kirsch2021stochastic}.

\citet{kirsch2019batchbald} explicitly model the interactions between data points by directly using the
\textit{joint} mutual information (JMI),
\begin{align}
    \label{eqn:jmi}
    R_{\text{BatchBALD}} = \mathbb{I}\left[y_{1:B}, \theta \mid x_{1:B}, \mathcal{D_{\text{train}}}\right].
\end{align} 
To deal with the intractability of the maximization problem, the authors propose a greedy approximation that is guaranteed to yield a batch for which the JMI is larger than $(1 - \frac{1}{e})$ times the optimal JMI, because of sub-modularity of the JMI set function. 
Greedy strategies for maximizing JMI in the regression setting have also been proposed \cite{holzmuller2022framework, wang2021beyond}.
Related to the JMI \cite{kirschunifying}, Fisher Information has also been explored as a batch acquisition objective \cite{ash2021gone}.
Other AL strategies such as Coresets \cite{seneractive} or LCMD \cite{holzmuller2022framework}, which use distance-based metrics to ensure diversity in the batch, also use greedy approximations to find a batch with high reward. 
In addition to the approximation, another major drawback of BatchBALD, and other batch-aware methods such as BADGE~\citep{ash2019deep}, is their high computational cost.

\subsection{Generative Flow Networks}\label{sec:gflownet}
Generative Flow Networks~\citep[GFlowNets; GFNs;][]{bengio2021flow,bengio2021gflownet} are probabilistic models over discrete sample spaces with a compositional structure. GFNs are stochastic sequential samplers that aim to generate objects from a target distribution, which is given by its unnormalized probability mass function $R$, also referred to as the {\em reward} function. 

The sample space, denoted $\mathcal{S}^f$, is the subset of terminal nodes (i.e. have no outgoing edges) of the vertices $\mathcal{S}$ of a directed acyclic graph (DAG) $\mathcal= \{\mathcal{S}, \mathbb{A}\}$ with a special parentless state $s_0$ called the source state. $\mathcal{S}$ consists of partially constructed objects, $s_0$ being an empty object, and $\mathbb{A}$ corresponds to actions that can be taken at each of these states. Complete objects $s_f\in\mathcal{S}^f$ are sampled by following a complete trajectory $\tau$ starting at $s_0$ and terminating at $s_f$. 

GFN training objectives allow the learning of a \emph{policy} $P_F(s' \mid s)$ along the edges of the DAG with the goal of making the marginal likelihood of sampling $s_f$ proportional to the reward $R(s_f)$. The parameters of $P_F$ are sequentially updated using the gradient of one of the losses applied to trajectories (or parts of trajectories) sampled from the trajectory distribution induced by $P_F$ (or some exploratory distribution, such as a tempered $P_F$). The various GFN losses in common use make use of a parametric \emph{backward policy} $P_B(s \mid s')$, specifying distributions over parents of the states in the DAG, and optionally a \emph{state flow} function $F(s)$ \cite{malkin2022trajectory}. In this work, we use the Subtrajectory Balance objective~\citep{madan2022learning} which provides advantages in training stability. We also leverage the forward-looking parametrization~\citep{pan2023better}, which uses the stepwise gain in a proxy log-reward computed at intermediate states to improve credit assignment. 

The parametric objects learned by GFNs, as well as their rewards, can be conditioned on instance-specific information, in our case a training set and a pool set, enabling generalization to conditioning data not seen in training~\citep{jain2022multi,zhang2023robust}.

\section{Methods}
\label{sec:methods}

\subsection{BatchGFN: A Sampler for Batches of Data}
\label{sec:method-BatchGFN}

We propose BatchGFN; a parameterized sampler for batch AL which uses a GFN trained to sample informative query batches of data to label. Prior work on GFNs in the context of active learning does not consider the pool-based setting and instead leverages the GFNs to sample individual candidates that comprise the batch, one at a time~\citep{jain2022biological}. 

Instead, we use a GFN to construct a batch (a set) of candidates from a pool in a single trajectory. As illustrated in Figure~\ref{fig:BatchGFN}, $P_F$ generates a batch of size $B$ through a sequence of steps, each consisting of adding an element from $\mathcal{D}_{\text{pool}}$ to the partially constructed batch. Note that the GFN is conditioned on $\mathcal{D}_{\text{train}}$. Each sampled batch $\{x_1, \dots, x_B\}$ is scored with a reward function $R(\cdot | \mathcal{D}_{\text{train}})$ which quantifies the utility of acquiring the batch. In this work we use the JMI (\ref{eqn:jmi}) to account for overlap in information between points in the batch. We can however, in principle, use \textit{any} heuristic that provides a scalar reward to a given batch.

With sufficient capacity, the BatchGFN converges to the true reward distribution in the limit of infinite training trajectories~\citep{bengio2021flow}. Therefore, once appropriately trained, BatchGFN can be used to sample batches for AL from the \textit{true batch objective} (\ref{eqn:jmi}) efficiently, without resorting to greedy, or stochastic top-$B$ approximations. BatchGFN has a time complexity of $\mathcal{O}(B)$ for sampling a batch, requiring only $B$ forward passes of $P_{F}$. This is cheap compared to, for example, BatchBALD which requires computing joint entropies for all points in the pool which can be particularly expensive when using Monte-Carlo (MC) samples \cite{kirsch2019batchbald}.

\begin{figure}
    \centering
    \includegraphics[width=\linewidth]{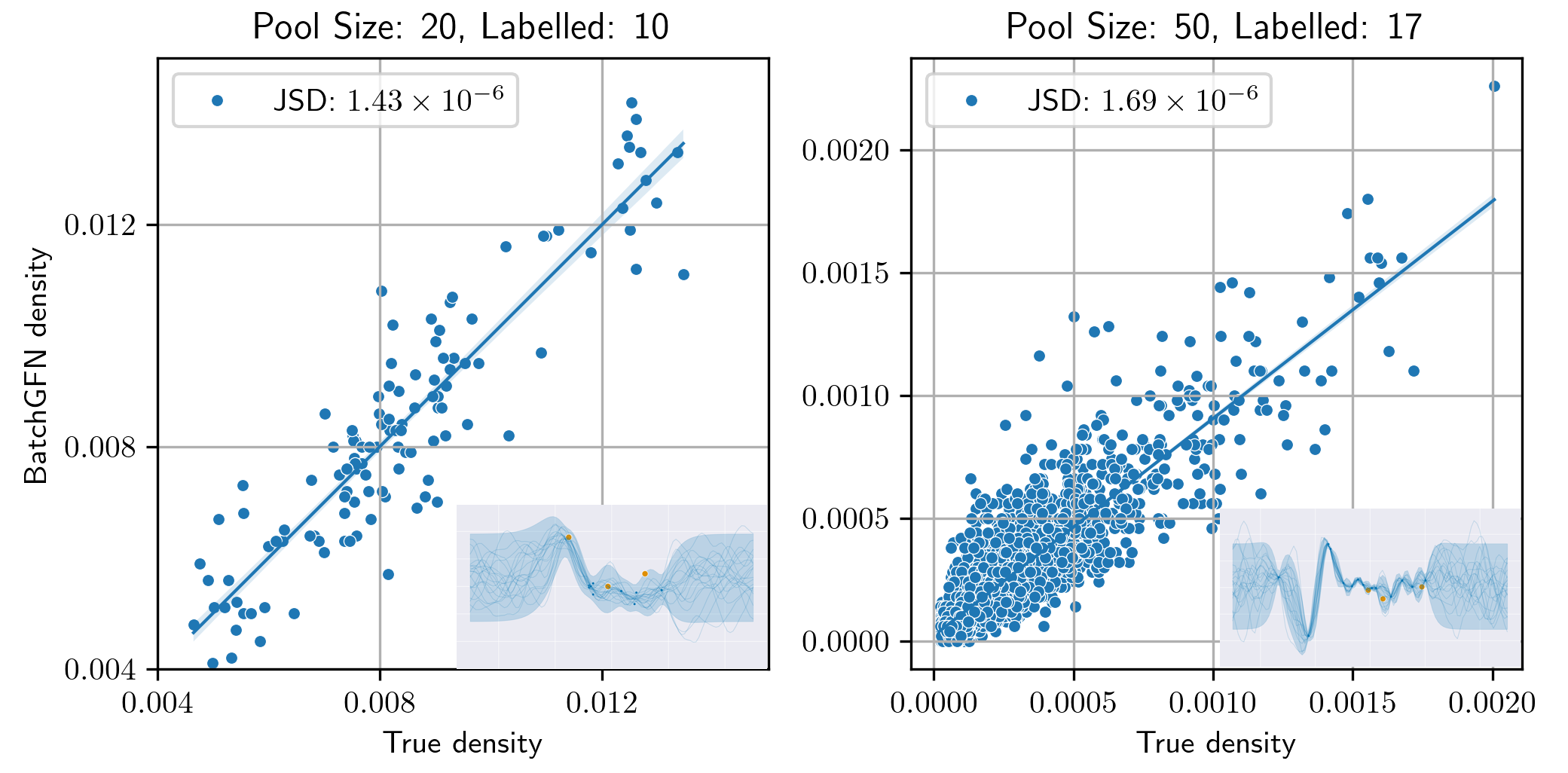}\vspace{-1em}
    \caption{Density parity plots comparing the true reward density for query batches against the empirical distribution of batches sampled from the BatchGFN. Regression lines are also shown. A perfect fit would have all points lying on $y=x$. The inlays shows the current model and labelled/queried data points in blue/orange respectively, and the Jenson-Shannon divergence between the two distributions.
    }
    \label{fig:density_plot}
\end{figure}

\subsection{Amortizing Training Across Acquisition Steps}
\label{sec:method-amort}

The reward distribution over possible query batches changes after each acquisition as the model is trained on the newly labelled data. 
Na\"{i}vely, the GFN therefore needs to be retrained to fit the new distribution at each AL step, which can be expensive. 
In theory, we could train models on samples from a distribution of $\mathcal{D}_{\text{train}}$ and and use these as examples to train an amortized $P_F$. 

Instead, we draw on ideas from GP ``fantasization'' literature \cite{hennig2012entropy, jiang2020efficient, maddox2021conditioning} to ``lookahead'' to possible future reward distributions after acquisition. We use samples from the current model to hallucinate labels for the next chosen query batch, add these to the training set, retrain the model, and then train the sampler using the new reward function and training set. This greatly restricts the space of conditional reward distributions required to be modelled and thus the computational expense required for training. See Appendix \ref{sec:app-lookahead} for further details on lookahead training.

\begin{figure}
    \centering
    \includegraphics[width=\linewidth]{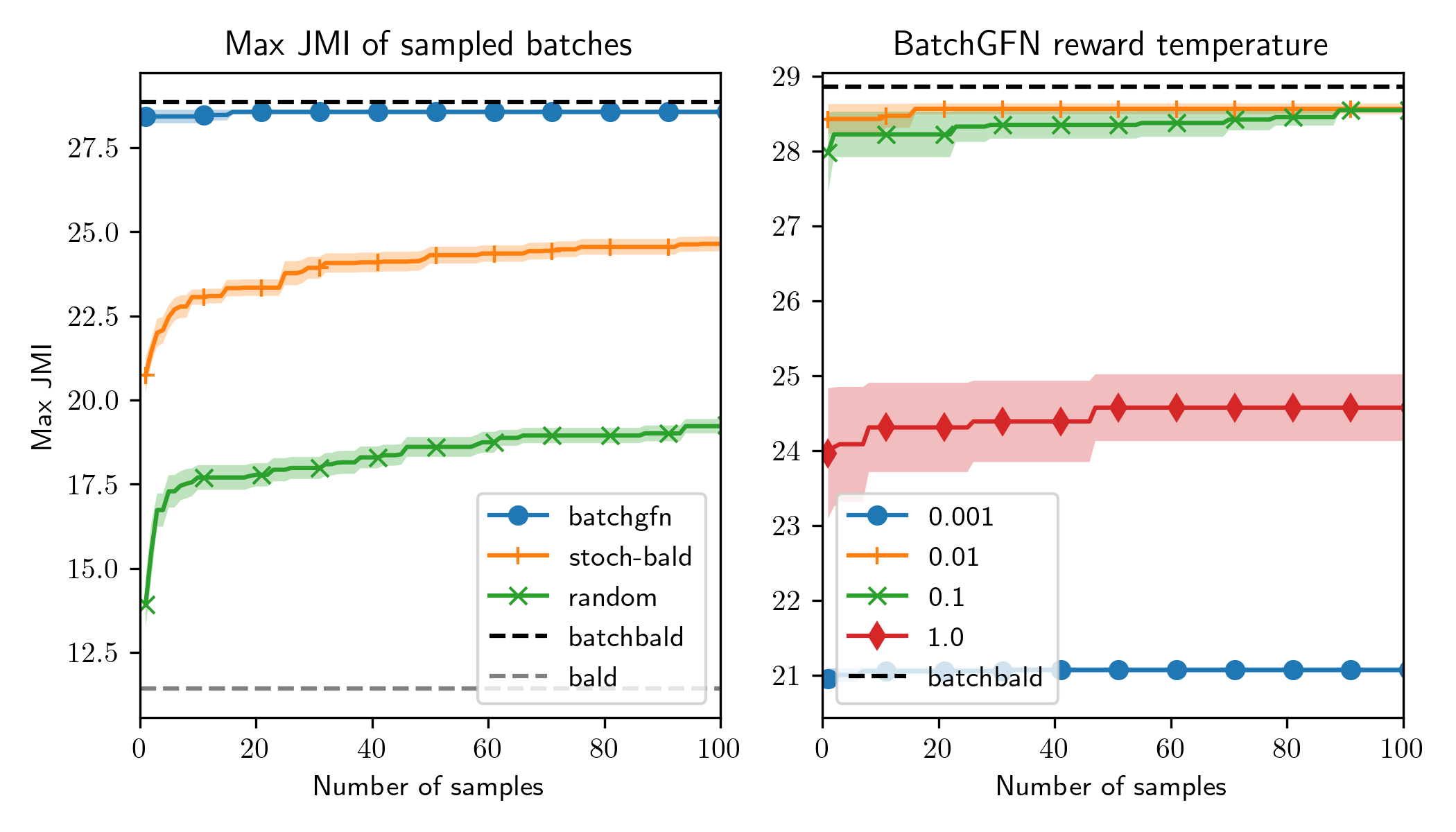}\vspace{-1em}
    \caption{Joint mutual information of sampled query batches from BatchGFN compared to baselines for the 1D regression task with pool size 2000, seed size 10, query size 10, and $T=0.01$. Uncertainty bars show the standard error over 10 sampled runs. 
    The stochastic-BALD baseline samples from a distribution of single-point BALD scores. The right plot compares BatchGFN sampling at different reward temperatures. We note that very low temperatures cause training instabilities. Conditioning the policy on the temperature may alleviate this behaviour \cite{zhang2023robust}.} 
    \label{fig:jmi_plot}
\end{figure}

\subsection{Implementation Details}
\label{sec:method-imp}

\paragraph{Batch Reward Function and Active Learning Model} 
In the experiments shown here we chose to use the JMI (\ref{eqn:jmi}) as the measure of utility for the constructed batches. We use exact inference Gaussian processes (GPs) as the model for AL. This allows us to use a closed form solution for the JMI between batch labels $y_{1:B}$ and the model $f$,
\begin{align}
    \label{eqn:gp-mi}
    \mathbb{I}\left(y_{1:B} ; f\right) = \mathbb{I}\left(y_{1:B} ; f_{1:B}\right)=\frac{1}{2} \log \left|\boldsymbol{I}+\sigma^{-2} \boldsymbol{K}_{1:B}\right|
\end{align}
where $\boldsymbol{K}_{1:B}=\left[k\left({x}, {x}^{\prime}\right)\right] {x}, {x}^{\prime} \in {x}_{1:B}$ is the covariance matrix for the batch, $\sigma^2$ is the GP observation noise variance, and $\boldsymbol{I}$ is the identity matrix \cite{holzmuller2022framework, srinivas2010gaussian}. 
For AL we would like to preferentially sample batches with high JMI. Thus in practice, 
we modify the reward function to $R=\exp(\mathbb{I}\left(y_{1:B} ; f\right)/T)$, which includes a temperature parameter $T$ that enables sampling from a more peaky reward distribution to focus on the modes.

\paragraph{Policy Network Architecture and Training} Note that our framework is agnostic to policy parametrization. For experiments presented here we use a simple set-invariant architecture which is conditioned on the current batch and the training data (Appendix \ref{sec:app-models}). 
We use the Subtrajectory Balance objective with forward-looking parametrization which exploits the submodularity of the JMI objective for intermediate rewards. During training, we encourage state exploration by sampling from an $\epsilon$-random policy
~\citep{bengio2021flow}\footnote{Our code is available at \url{https://github.com/s-a-malik/batchgfn}}.

\section{Experiments}
\label{sec:results}

First we show through a toy example that the BatchGFN converges to a policy that samples batches of points proportional to their reward as expected. Then we show that by tuning the reward temperature, we can efficiently sample highly informative batches for AL. Finally, we present early results into amortizing training across acquisition steps.

\subsection{Sampling Proportional to the Batch Reward}
\label{sec:exp-syn-reg}

We consider a toy 1D regression task to verify behaviour. See Appendix \ref{sec:app-data} for details on the dataset.
For small pool and query sizes, it is possible to exhaustively evaluate the reward for every possible query batch. We compare the true reward distribution to the empirical distribution of batches generated by the trained BatchGFN (with $T=1$) in Figure \ref{fig:density_plot}. We see that the BatchGFN samples batches approximately proportional to their batch reward as expected.

\begin{figure}
    \centering
    \includegraphics[width=\linewidth]{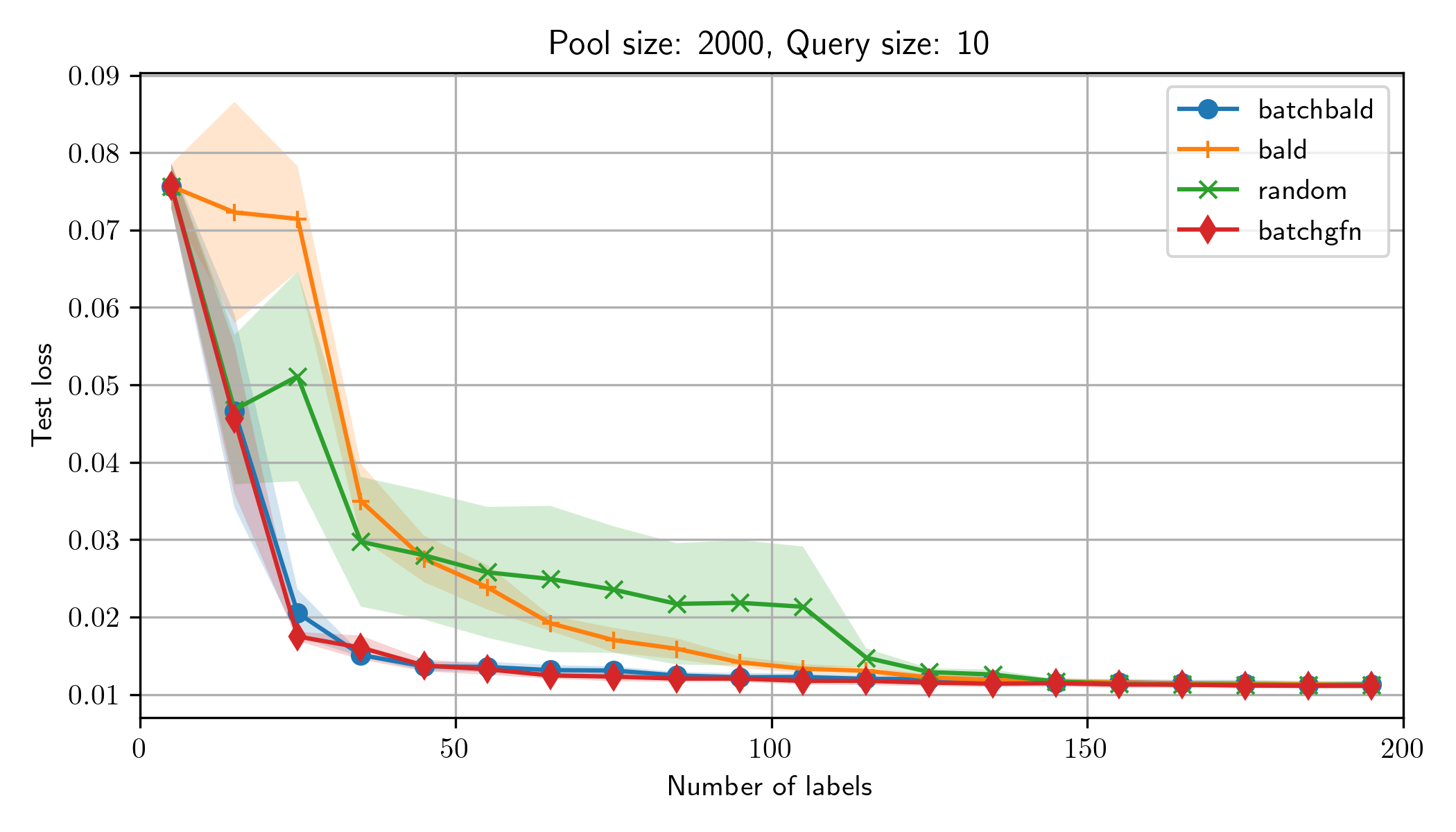}\vspace{-1em}
    \caption{Test loss as function of number of labelled examples for different active learning strategies. Active learning with BatchGFN is on par with BatchBALD. Uncertainty bars show the standard error over 5 runs with different seed sets.}
    \label{fig:al_reg}
\end{figure}

\subsection{Sampling High Joint Mutual Information Batches}
\label{sec:exp-jmi-sample}

In practice, we would like to sample high JMI batches for applications such as AL. Figure \ref{fig:jmi_plot} shows the JMI of sampled batches from the BatchGFN at different reward temperatures for the 1D regression task. We find that by decreasing the temperature we are able to sample higher JMI batches with greater sample efficiency compared to other stochastic approaches. The batches sampled are on par with BatchBALD while being less computationally expensive\footnote{We note that the speed-up in run-time compared to BatchBALD using the exact GP JMI (\ref{eqn:gp-mi}) in the current setting is modest ($\sim 10\%$ faster for a single sample from BatchGFN). However, BatchBALD is orders of magnitude more expensive when using MC sampling, whereas BatchGFN inference will be unaffected.}.

\subsection{Active Learning with BatchGFN}
\label{sec:exp-al}

We have shown that BatchGFN can sample high JMI batches effectively. Now we assess how effective these are for AL. Figure \ref{fig:al_reg} shows the test loss of the model at the toy regression task using queries sampled from BatchGFN. We sample 20 trajectories from the BatchGFN at inference and choose the batch with the highest reward. We find that BatchGFN is on par with BatchBALD and significantly outperforms BALD and random acquisition. Appendix \ref{sec:app-acq-plots} contains example acquisition plots.

\subsection{Amortizing Across Acquisition Steps}
\label{sec:exp-amort}

\begin{figure}
    \centering
    \includegraphics[width=\linewidth]{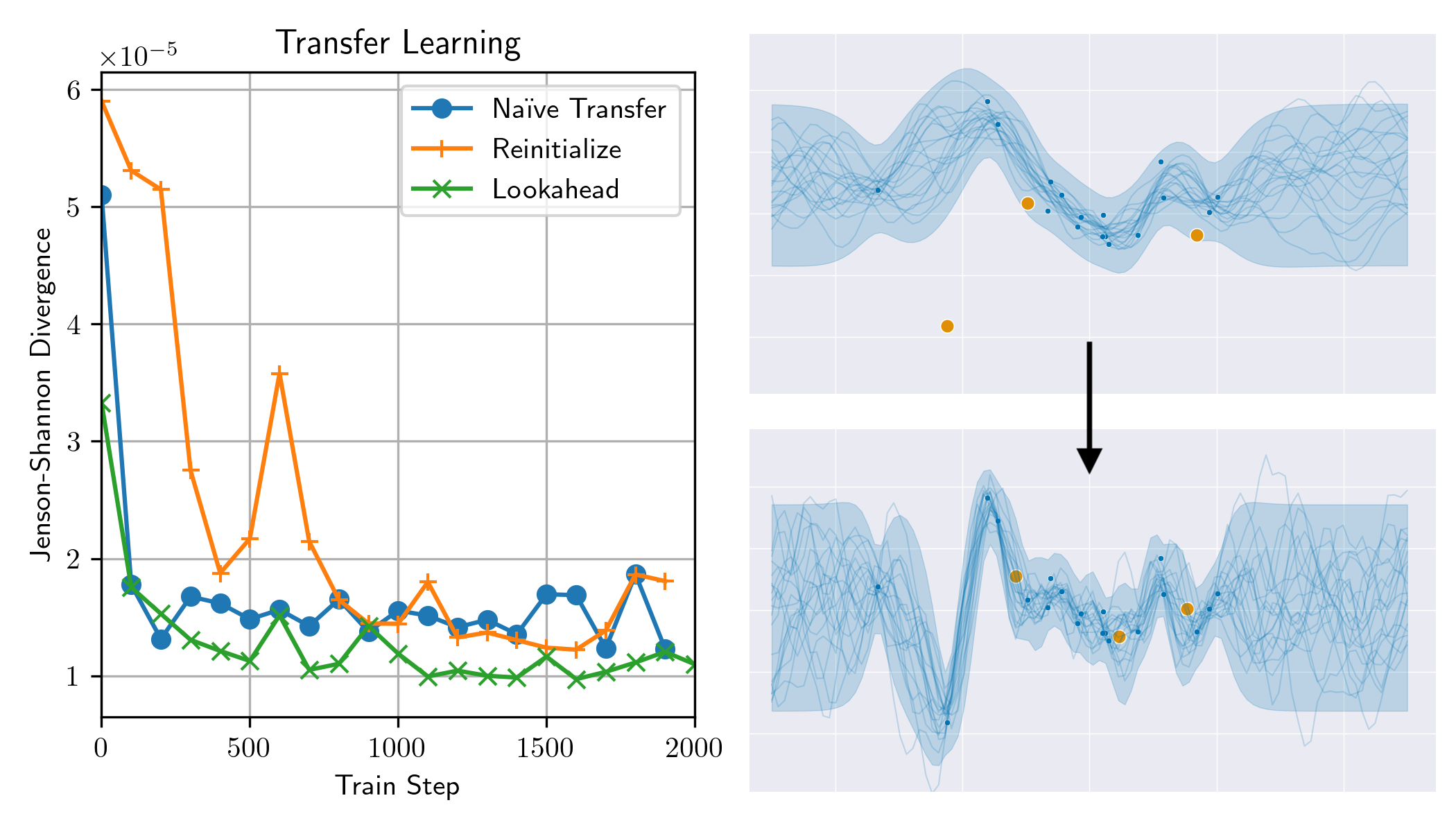}\vspace{-1em}
    \caption{Training curve plot showing the Jenson-Shannon divergence between the true reward distribution and the empirical BatchGFN distribution when transferring from one acquisition step to the next (shown on the right). We train on 10 hallucinated samples of the query batch to be labelled.}
    \label{fig:transfer-jsd}
\end{figure}

Retraining the BatchGFN between acquisition steps is impractical for real-world usage. 
As discussed in Section \ref{sec:method-amort}, we can amortize training over acquisition steps by conditioning the policy on the training data. In Figure \ref{fig:transfer-jsd} we compare how quickly the BatchGFN adapts to the new conditional reward distribution after an acquisition on a small pool/query size such that we can compute the true distribution (Section \ref{sec:exp-syn-reg}). 
We find that lookahead training significantly decreases the number of training steps required to fit to the new distribution compared to reinitializing the policy network. 
Na\"{i}vely continuing training from the previous reward distribution also works well but starts at higher divergence as it is overfit to the previous distribution, making it unsuitable for zero-shot transfer, and potentially unstable for few-shot transfer.  
This shows promise for amortizing BatchGFN training over multiple acquisition steps and for larger tasks.

\section{Discussion}
\label{sec:disc}

\paragraph{Contributions} 
In this work we developed BatchGFN---a novel method based on GFNs to sample batches of data points proportional to an arbitrary batch reward function. We have shown that this method can be used to efficiently sample high JMI batches for AL tasks in toy regression problems and has performance on par with more expensive batch construction methods, without resorting to greedy approximations. 

\paragraph{Further Work}
In Section \ref{sec:exp-amort} we showed that it is possible to amortize training. However for practical usage, we need to show that this is possible over longer acquisition step horizons, larger pool sets, and higher dimensional data. To do this, we can investigate the following: 1) Architecturally, we can improve the representational ability of the policy network by considering attention between data points~\cite{kossen2021self}. 2) To scale to larger pool sizes, we could incorporate perceiver-like bottlenecks~\cite{jaegle2021perceiver}. 3) To enable training on tasks where a cheap JMI estimate is not available, we should investigate alternative batch heuristics to train the GFN efficiently.

\section*{Acknowledgements}

We would like to thank Panagiotis Tigas and Andreas Kirsch for helpful discussions about the project. SM acknowledges funding from the EPSRC Centre for Doctoral Training in Autonomous Intelligent Machines and Systems (Grant No: EP/S024050/1). YG acknowledges funding from the Turing Fellowship (Grant No. EP/V030302/1). MJ, NM, TD, and YB acknowledge financial support from IBM, Samsung, Microsoft, and Google.

\bibliography{main}

\begin{thebibliography}{29}
\providecommand{\natexlab}[1]{#1}
\providecommand{\url}[1]{\texttt{#1}}
\expandafter\ifx\csname urlstyle\endcsname\relax
  \providecommand{\doi}[1]{doi: #1}\else
  \providecommand{\doi}{doi: \begingroup \urlstyle{rm}\Url}\fi

\bibitem[Ash et~al.(2021)Ash, Goel, Krishnamurthy, and Kakade]{ash2021gone}
Ash, J., Goel, S., Krishnamurthy, A., and Kakade, S.
\newblock Gone fishing: Neural active learning with fisher embeddings.
\newblock \emph{Advances in Neural Information Processing Systems},
  34:\penalty0 8927--8939, 2021.

\bibitem[Ash et~al.(2019)Ash, Zhang, Krishnamurthy, Langford, and
  Agarwal]{ash2019deep}
Ash, J.~T., Zhang, C., Krishnamurthy, A., Langford, J., and Agarwal, A.
\newblock Deep batch active learning by diverse, uncertain gradient lower
  bounds.
\newblock \emph{arXiv preprint arXiv:1906.03671}, 2019.

\bibitem[Bengio et~al.(2021{\natexlab{a}})Bengio, Jain, Korablyov, Precup, and
  Bengio]{bengio2021flow}
Bengio, E., Jain, M., Korablyov, M., Precup, D., and Bengio, Y.
\newblock Flow network based generative models for non-iterative diverse
  candidate generation.
\newblock In Beygelzimer, A., Dauphin, Y., Liang, P., and Vaughan, J.~W.
  (eds.), \emph{Advances in Neural Information Processing Systems},
  2021{\natexlab{a}}.
\newblock URL \url{https://openreview.net/forum?id=Arn2E4IRjEB}.

\bibitem[Bengio et~al.(2021{\natexlab{b}})Bengio, Deleu, Hu, Lahlou, Tiwari,
  and Bengio]{bengio2021gflownet}
Bengio, Y., Deleu, T., Hu, E.~J., Lahlou, S., Tiwari, M., and Bengio, E.
\newblock Gflownet foundations, 2021{\natexlab{b}}.

\bibitem[Gal et~al.(2017)Gal, Islam, and Ghahramani]{gal2017deep}
Gal, Y., Islam, R., and Ghahramani, Z.
\newblock Deep bayesian active learning with image data.
\newblock In \emph{International Conference on Machine Learning}, pp.\
  1183--1192. PMLR, 2017.

\bibitem[Gardner et~al.(2018)Gardner, Pleiss, Weinberger, Bindel, and
  Wilson]{gardner2018gpytorch}
Gardner, J., Pleiss, G., Weinberger, K.~Q., Bindel, D., and Wilson, A.~G.
\newblock Gpytorch: Blackbox matrix-matrix gaussian process inference with gpu
  acceleration.
\newblock \emph{Advances in neural information processing systems}, 31, 2018.

\bibitem[Hennig \& Schuler(2012)Hennig and Schuler]{hennig2012entropy}
Hennig, P. and Schuler, C.~J.
\newblock Entropy search for information-efficient global optimization.
\newblock \emph{Journal of Machine Learning Research}, 13\penalty0 (6), 2012.

\bibitem[Holzm{\"u}ller et~al.(2022)Holzm{\"u}ller, Zaverkin, K{\"a}stner, and
  Steinwart]{holzmuller2022framework}
Holzm{\"u}ller, D., Zaverkin, V., K{\"a}stner, J., and Steinwart, I.
\newblock A framework and benchmark for deep batch active learning for
  regression.
\newblock \emph{arXiv preprint arXiv:2203.09410}, 2022.

\bibitem[Houlsby et~al.(2011)Houlsby, Huszár, Ghahramani, and
  Lengyel]{houlsby2011bayesian}
Houlsby, N., Huszár, F., Ghahramani, Z., and Lengyel, M.
\newblock Bayesian active learning for classification and preference learning.
\newblock \emph{arXiv preprint arXiv: Arxiv-1112.5745}, 2011.

\bibitem[Jaegle et~al.(2021)Jaegle, Gimeno, Brock, Vinyals, Zisserman, and
  Carreira]{jaegle2021perceiver}
Jaegle, A., Gimeno, F., Brock, A., Vinyals, O., Zisserman, A., and Carreira, J.
\newblock Perceiver: General perception with iterative attention.
\newblock In \emph{International conference on machine learning}, pp.\
  4651--4664. PMLR, 2021.

\bibitem[Jain et~al.(2022{\natexlab{a}})Jain, Bengio, Hernandez-Garcia,
  Rector-Brooks, Dossou, Ekbote, Fu, Zhang, Kilgour, Zhang,
  et~al.]{jain2022biological}
Jain, M., Bengio, E., Hernandez-Garcia, A., Rector-Brooks, J., Dossou, B.~F.,
  Ekbote, C.~A., Fu, J., Zhang, T., Kilgour, M., Zhang, D., et~al.
\newblock Biological sequence design with gflownets.
\newblock In \emph{International Conference on Machine Learning}, pp.\
  9786--9801. PMLR, 2022{\natexlab{a}}.

\bibitem[Jain et~al.(2022{\natexlab{b}})Jain, Raparthy, Hernandez-Garcia,
  Rector-Brooks, Bengio, Miret, and Bengio]{jain2022multi}
Jain, M., Raparthy, S.~C., Hernandez-Garcia, A., Rector-Brooks, J., Bengio, Y.,
  Miret, S., and Bengio, E.
\newblock Multi-objective gflownets.
\newblock \emph{arXiv preprint arXiv:2210.12765}, 2022{\natexlab{b}}.

\bibitem[Jiang et~al.(2020)Jiang, Jiang, Balandat, Karrer, Gardner, and
  Garnett]{jiang2020efficient}
Jiang, S., Jiang, D., Balandat, M., Karrer, B., Gardner, J., and Garnett, R.
\newblock Efficient nonmyopic bayesian optimization via one-shot multi-step
  trees.
\newblock \emph{Advances in Neural Information Processing Systems},
  33:\penalty0 18039--18049, 2020.

\bibitem[Kingma \& Ba(2014)Kingma and Ba]{kingma2014adam}
Kingma, D.~P. and Ba, J.
\newblock Adam: A method for stochastic optimization.
\newblock \emph{arXiv preprint arXiv:1412.6980}, 2014.

\bibitem[Kirsch \& Gal(2022)Kirsch and Gal]{kirschunifying}
Kirsch, A. and Gal, Y.
\newblock Unifying approaches in active learning and active sampling via fisher
  information and information-theoretic quantities.
\newblock \emph{Transactions on Machine Learning Research}, 2022.

\bibitem[Kirsch et~al.(2019)Kirsch, Van~Amersfoort, and
  Gal]{kirsch2019batchbald}
Kirsch, A., Van~Amersfoort, J., and Gal, Y.
\newblock Batchbald: Efficient and diverse batch acquisition for deep bayesian
  active learning.
\newblock \emph{Advances in neural information processing systems}, 32, 2019.

\bibitem[Kirsch et~al.(2021)Kirsch, Farquhar, Atighehchian, Jesson,
  Branchaud-Charron, and Gal]{kirsch2021stochastic}
Kirsch, A., Farquhar, S., Atighehchian, P., Jesson, A., Branchaud-Charron, F.,
  and Gal, Y.
\newblock Stochastic batch acquisition for deep active learning.
\newblock \emph{arXiv preprint arXiv:2106.12059}, 2021.

\bibitem[Kossen et~al.(2021)Kossen, Band, Lyle, Gomez, Rainforth, and
  Gal]{kossen2021self}
Kossen, J., Band, N., Lyle, C., Gomez, A.~N., Rainforth, T., and Gal, Y.
\newblock Self-attention between datapoints: Going beyond individual
  input-output pairs in deep learning.
\newblock \emph{Advances in Neural Information Processing Systems},
  34:\penalty0 28742--28756, 2021.

\bibitem[Lahlou et~al.(2023)Lahlou, Viviano, and Schmidt]{lahlou2023torchgfn}
Lahlou, S., Viviano, J.~D., and Schmidt, V.
\newblock torchgfn: A pytorch gflownet library.
\newblock \emph{arXiv preprint arXiv: 2305.14594}, 2023.

\bibitem[Lewis(1995)]{lewis1995sequential}
Lewis, D.~D.
\newblock A sequential algorithm for training text classifiers: Corrigendum and
  additional data.
\newblock In \emph{Acm Sigir Forum}, volume~29, pp.\  13--19. ACM New York, NY,
  USA, 1995.

\bibitem[Madan et~al.(2023, to appear.)Madan, Rector-Brooks, Korablyov, Bengio,
  Jain, Nica, Bosc, Bengio, and Malkin]{madan2022learning}
Madan, K., Rector-Brooks, J., Korablyov, M., Bengio, E., Jain, M., Nica, A.,
  Bosc, T., Bengio, Y., and Malkin, N.
\newblock Learning gflownets from partial episodes for improved convergence and
  stability.
\newblock \emph{International Conference on Machine Learning (ICML)}, 2023, to
  appear.

\bibitem[Maddox et~al.(2021)Maddox, Stanton, and
  Wilson]{maddox2021conditioning}
Maddox, W.~J., Stanton, S., and Wilson, A.~G.
\newblock Conditioning sparse variational gaussian processes for online
  decision-making.
\newblock \emph{Advances in Neural Information Processing Systems},
  34:\penalty0 6365--6379, 2021.

\bibitem[Malkin et~al.(2022)Malkin, Jain, Bengio, Sun, and
  Bengio]{malkin2022trajectory}
Malkin, N., Jain, M., Bengio, E., Sun, C., and Bengio, Y.
\newblock Trajectory balance: Improved credit assignment in gflownets.
\newblock \emph{Neural Information Processing Systems (NeurIPS)}, 2022.

\bibitem[Pan et~al.(2023, to appear.)Pan, Malkin, Zhang, and
  Bengio]{pan2023better}
Pan, L., Malkin, N., Zhang, D., and Bengio, Y.
\newblock Better training of gflownets with local credit and incomplete
  trajectories.
\newblock \emph{International Conference on Machine Learning (ICML)}, 2023, to
  appear.

\bibitem[Sener \& Savarese(2018)Sener and Savarese]{seneractive}
Sener, O. and Savarese, S.
\newblock Active learning for convolutional neural networks: A core-set
  approach.
\newblock In \emph{International Conference on Learning Representations}, 2018.

\bibitem[Settles(2009)]{settles2009active}
Settles, B.
\newblock Active learning literature survey.
\newblock 2009.

\bibitem[Srinivas et~al.(2010)Srinivas, Krause, Kakade, and
  Seeger]{srinivas2010gaussian}
Srinivas, N., Krause, A., Kakade, S., and Seeger, M.
\newblock Gaussian process optimization in the bandit setting: no regret and
  experimental design.
\newblock In \emph{Proceedings of the 27th International Conference on
  International Conference on Machine Learning}, pp.\  1015--1022, 2010.

\bibitem[Wang et~al.(2021)Wang, Sun, and Grosse]{wang2021beyond}
Wang, C., Sun, S., and Grosse, R.
\newblock Beyond marginal uncertainty: How accurately can bayesian regression
  models estimate posterior predictive correlations?
\newblock In \emph{International Conference on Artificial Intelligence and
  Statistics}, pp.\  2476--2484. PMLR, 2021.

\bibitem[Zhang et~al.(2023)Zhang, Rainone, Peschl, and
  Bondesan]{zhang2023robust}
Zhang, D.~W., Rainone, C., Peschl, M., and Bondesan, R.
\newblock Robust scheduling with gflownets.
\newblock \emph{arXiv preprint arXiv:2302.05446}, 2023.

\end{thebibliography}
\bibliographystyle{icml2023}

\newpage
\appendix
\onecolumn

\section{Further Experimental Details}
\label{sec:app-implementation}

\subsection{Data}
\label{sec:app-data}

\begin{figure}[h]
    \centering
    \includegraphics[width=0.5\linewidth]{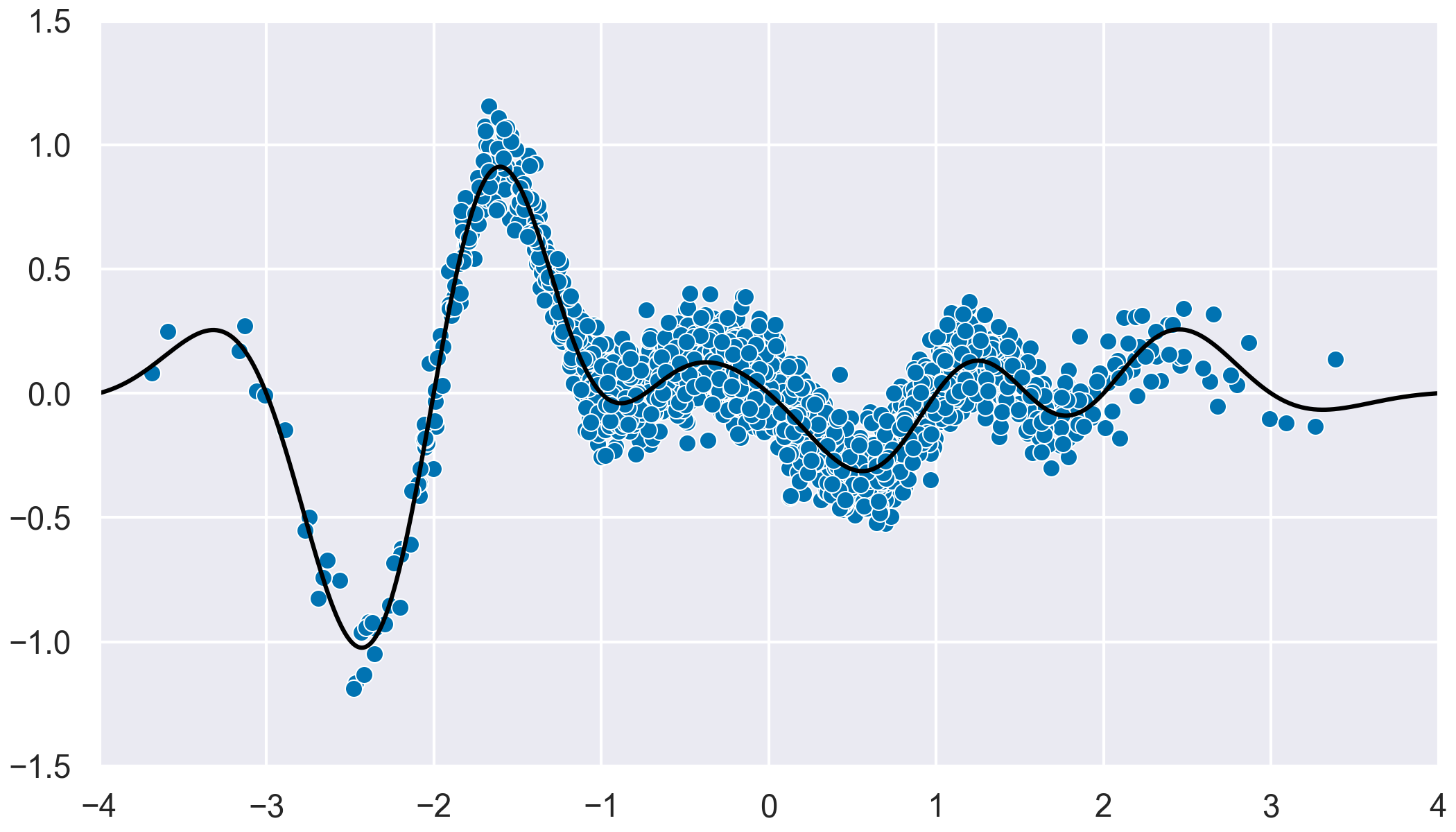}
    \caption{Toy example 1D regression data.}
    \label{fig:reg-data}
\end{figure}

\paragraph{Toy Regression Task}
The toy regression task used for experiments in Section \ref{sec:results} is shown in Figure \ref{fig:reg-data}. The pool and test data was generated from the function $f(x) = (-0.6667 -0.6012x -1.0172x^2-0.7687x^3 +1.4680x^5 -0.1678x^6)\sin(\pi x)\exp(-0.5x^2) + N(0, 0.1)$, and $x\sim N(0,1)$ where $N$ is the normal distribution. The test set was fixed at 1000 points and generated from a different random seed. Seed datasets for active learning experiments were chosen randomly from the pool set.

\subsection{Models}
\label{sec:app-models}

\begin{figure}[h]
    \centering
    \includegraphics[width=0.75\linewidth]{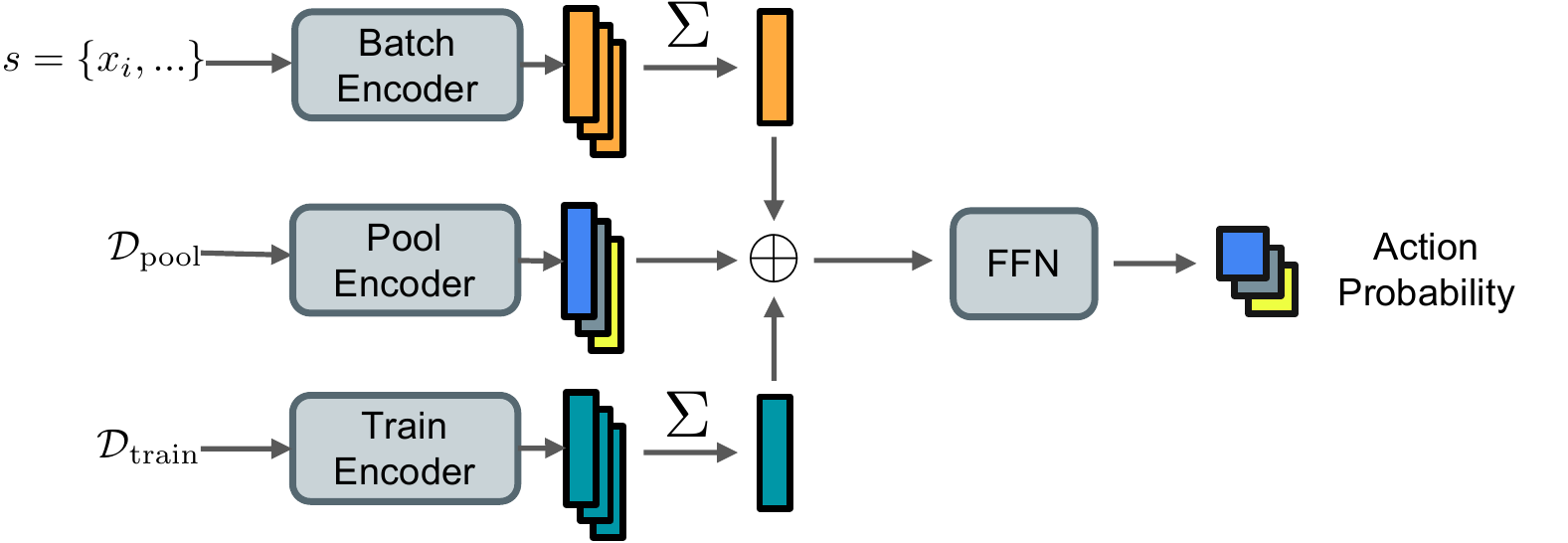}
    \caption{Policy network architecture. For non-transfer experiments we do not include the train encoder. The outputs of the policy network are the probabilities for adding each pool point to the current state, i.e. the query batch being built.}
    \label{fig:policy-arch}
\end{figure}

The policy network architecture is shown in Figure \ref{fig:policy-arch}. We use feed-forward neural network (FFN) encoders to encode each pool point and points in the current state. The embeddings of the points in the current state are summed (for set-invariance), and concatenated with each embedded pool point as a context vector. The combined embedding is fed through an additional FFN to output the probability of selecting each pool point. We share all parameters apart from output layers across $P_F$, $P_B$ and the log-state-flow network \cite{madan2022learning}. For amortization experiments, we concatenate an additional context vector of summed embeddings of the training set points and their labels. 

Hyperparameters for the policy network and the GP used for active learning are given in Table \ref{tab:hypers}. We use the default GPyTorch~\cite{gardner2018gpytorch} settings for implementing the GPs\footnote{Our code is available at \url{https://github.com/s-a-malik/batchgfn}. It relies on torchgfn~\citep{lahlou2023torchgfn} for implementing the GFlowNets.}. 

\begin{table}[h]
\caption{Hyperparameter configuration for active learning regression experiments.}
\begin{tabular}{@{}p{0.3\linewidth}p{0.65\linewidth}@{}}
\toprule
\textbf{Hyperparameter}  & \textbf{Value}    \\ \midrule \midrule
\textbf{BatchGFN Optimization}  &   \\ \midrule 
Loss       & SubTB (forward-looking) \cite{madan2022learning, malkin2022trajectory} \\
SubTB $\lambda$     &  0.9    \\
Training exploration $\epsilon$       &  0.1    \\
Reward Temperature $T$       &  0.1    \\
Optimizer          & Adam \cite{kingma2014adam}    \\     
Learning Rate   & 0.001      \\
Batch Size       & 8     \\
Training Iterations   & 5000    \\ \midrule \midrule
\textbf{BatchGFN Architecture}   &   \\ \midrule 
Hidden Layer Dimension          & 256       \\ 
Number of Encoder Hidden Layers         & 2     \\ 
Number of Batch Samples for Inference     & 20 
    \\ \midrule \midrule
\textbf{Lookahead Experiments}                                  &                                                                \\ \midrule 
Lookahead Samples   &    10  \\
Reward Temperature    &  0.1  \\
Seed Size   &  17  \\
Query Size   &  3  \\
Pool Size   &  50  \\
Pool sampling for lookahead  &   BatchGFN with training exploration  \\  \midrule \midrule
\textbf{Active Learning Model}                                  &                                                                \\ \midrule 
Model   &    Exact Gaussian Process  \\
Hyperparameter Training Epochs   &    1000  \\ 
Optimizer   &    Adam \cite{kingma2014adam}  \\ 
Learning Rate   &    0.1  \\ 
Kernel   &    Mat\'{e}rn  \\ 
\bottomrule 
 
\end{tabular}
\label{tab:hypers}
\end{table}

\subsection{Active Learning with BatchGFN}
\label{sec:app-lookahead}

Algorithm \ref{alg:lookahead} shows pseudocode for active learning with BatchGFN, including optional lookahead training where we hallucinate labels and train on possible future reward distributions. We are able to transfer to the new conditional reward distribution with fewer BatchGFN training iterations by using lookahead training (Section \ref{sec:exp-amort}).

\begin{algorithm}[h]
\caption{BatchGFN active learning with optional lookahead training.}
\label{alg:lookahead}
\begin{algorithmic}

\REQUIRE Query batch size: $B$, Seed set size: $B_0$, Pool dataset: $\mathcal{D}_{\text{pool}}$: $\{x_0, \ldots, x_N\}$
\REQUIRE Active learning model: $f$, BatchGFN: $g$, Reward Function: $R$, Lookahead samples: $L$
\STATE Randomly sample $\{x^*_0, \ldots, x^*_{B_0}\}$ from $\mathcal{D}_{pool}$
\STATE Label seed batch, $y^*_x \leftarrow$ Oracle($x$) $\forall x \in \{x^*_0, \ldots, x^*_{B_0}\}$
\STATE $\mathcal{D}_{\text{train}} \leftarrow \{(x^*_0, y^*_0), \ldots, (x^*_{B_0}, y^*_{B_0})\}$
\STATE $\mathcal{D}_{\text{pool}}\leftarrow \mathcal{D}_{\text{pool}} \setminus \{x^*_0, \ldots, x^*_{B_0}\}$

\WHILE{labelling budget not exhausted}
    \STATE Train $f$ on $\mathcal{D}_{\text{train}}$
    \STATE Train $g$ using $\mathcal{D}_{\text{train}}$, $\mathcal{D}_{\text{pool}}$, $f$, and $R$
    \STATE Sample batches from $g$
    \STATE batch $\leftarrow \max_R (\text{batches})$
    \IF{lookahead}
        \FOR{$i=1$ to $L$}
            \STATE Sample lookahead\_batch from $g$
            \STATE{Hallucinate labels $y'_x \leftarrow f.\text{sample}(x') \forall x' \in$ lookahead\_batch}
            \STATE $\mathcal{D}'_{\text{train}} \leftarrow \mathcal{D}_{\text{train}} \cup \{(x'_0, y'_0), \ldots, (x'_B, y'_B)\} $ 
            \STATE $\mathcal{D}'_{\text{pool}} \leftarrow \mathcal{D}_{\text{pool}} \setminus \{x'_0,\ldots, x'_{B}\}$
            \STATE Train $f$ on $\mathcal{D}'_{\text{train}}$
            \STATE Train $g$ using $\mathcal{D}'_{\text{train}}$, $\mathcal{D}'_{\text{pool}}$, $f$, and $R$
        \ENDFOR
    \ENDIF
    \STATE Get true labels $y^*_x \leftarrow$ Oracle($x$) $\forall x \in$ batch
    \STATE $\mathcal{D}_{\text{train}} \leftarrow \mathcal{D}_{\text{train}} \cup \{(x^*_0, y^*_0), \ldots, (x^*_B, y^*_B)\} $ 
    \STATE $\mathcal{D}_{\text{pool}} \leftarrow \mathcal{D}_{\text{pool}} \setminus \{x^*_0,\ldots, x^*_{B}\}$
\ENDWHILE
\end{algorithmic}
\end{algorithm}

\section{Further Results}
\label{sec:app-further-results}

\subsection{Acquisition Plots}
\label{sec:app-acq-plots}

\begin{figure}[h]
    \centering
    \includegraphics[width=\linewidth]{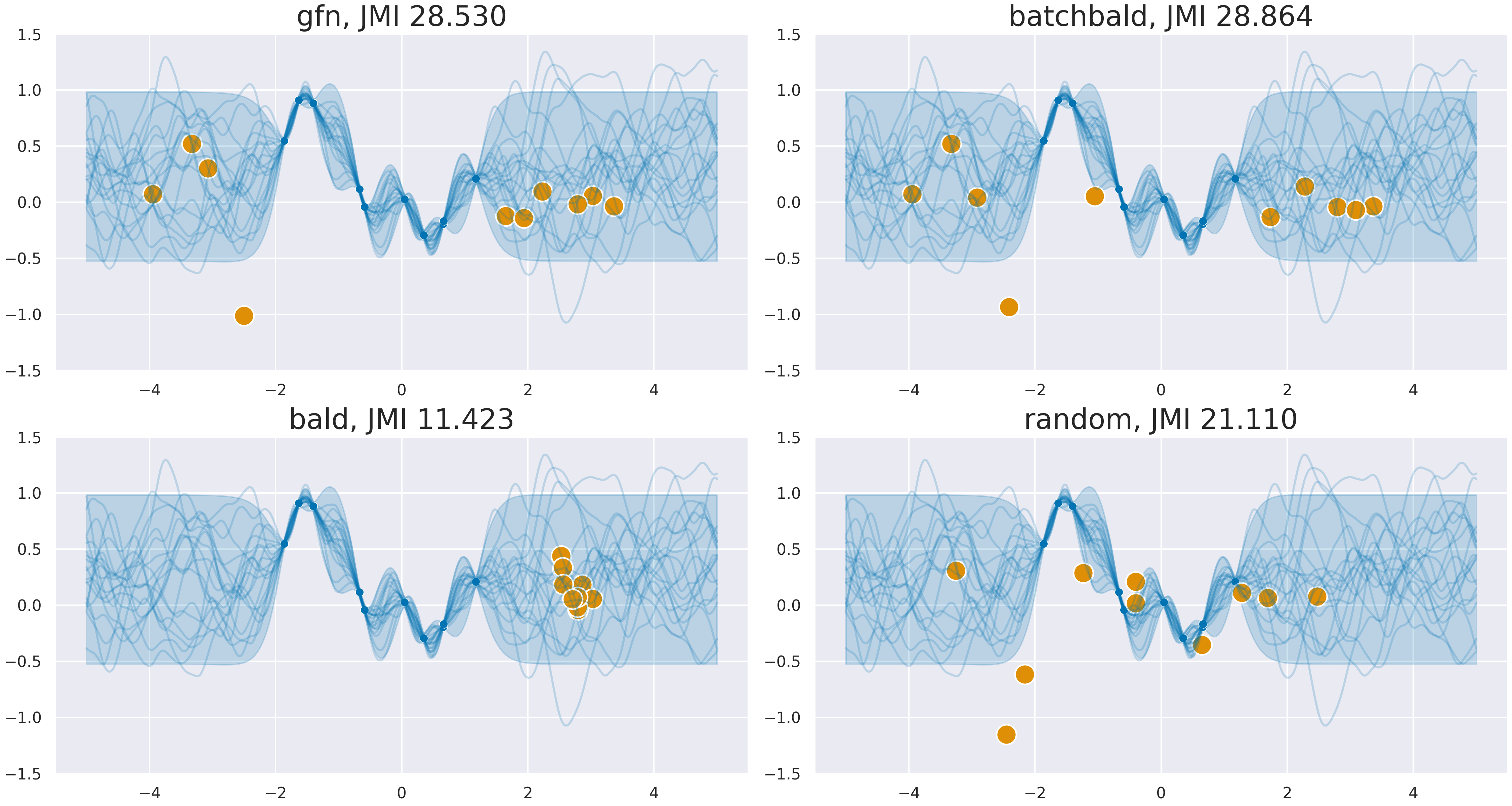}
    \caption{Acquisition plots for different active learning strategies. Pool size 2000, query size 10, training set size 10. Labelled/queried data points are shown in blue/orange respectively.}
    \label{fig:reg-acq}
\end{figure}

Figure \ref{fig:reg-acq} shows example acquisition plots comparing BatchGFN (gfn) acquisition to other baselines. BatchGFN acquires points that are diverse and uncertain like BatchBALD, whereas BALD acquires similar points. 

\subsection{Further Amortization Plots}

\begin{figure}[h]
    \centering
    \includegraphics[width=0.5\linewidth]{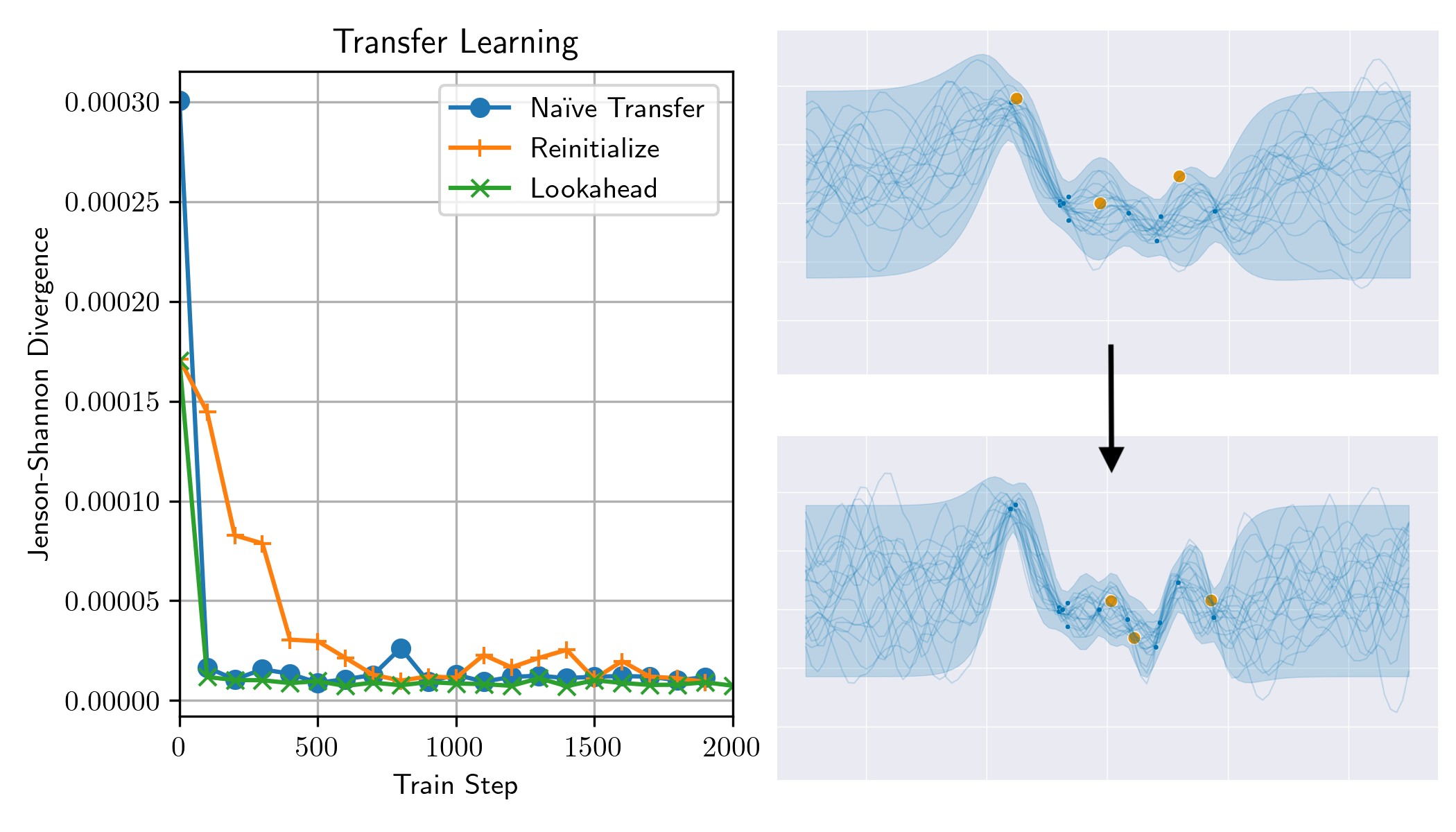}
    \caption{Training curve plot showing the Jenson-Shannon divergence between the true reward distribution and the empirical BatchGFN distribution when transferring from one acquisition step to the next (shown on the right). We use 10 lookahead samples.}
    \label{fig:transfer-jsd-20}
\end{figure}

We provide an additional example of transfer between AL steps on a smaller pool set of size 20 in Figure \ref{fig:transfer-jsd-20}. Lookahead training again allows for faster convergence to the true distribution.

\end{document}